\documentclass[sigconf, author]{acmart}
\usepackage{standalone}
\usepackage{color,soul}

\usepackage{algorithm}
\usepackage{algpseudocode}
\usepackage{subfig,tikz}
\tikzset{node_style/.style={circle,draw=black,fill=gray!20!}}
\tikzset{edge_style/.style={draw=black, thick}}
\tikzset{gexamples_node_style/.style={circle,draw=black,fill=gray!20!,scale=0.8}}

\algnewcommand\algorithmicforeach{\textbf{for each}}
\algdef{S}[FOR]{ForEach}[1]{\algorithmicforeach\ #1\ \algorithmicdo}
\usepackage{algpseudocode}
\algrenewcommand\textproc{}

\let\oldReturn\Return
\renewcommand{\Return}{\State\oldReturn}

\copyrightyear{2018} 
\acmYear{2018} 
\setcopyright{acmcopyright}
\acmConference[GECCO '18]{Genetic and Evolutionary Computation Conference}{July 15--19, 2018}{Kyoto, Japan}
\acmBooktitle{GECCO '18: Genetic and Evolutionary Computation Conference, July 15--19, 2018, Kyoto, Japan}
\acmPrice{15.00}
\acmDOI{10.1145/3205455.3205555}
\acmISBN{978-1-4503-5618-3/18/07}

\begin{document}
\title{Limited Evaluation Cooperative Co-evolutionary Differential\\ Evolution for Large-scale Neuroevolution}

\author{Anil Yaman}
\affiliation{%
  \institution{Eindhoven University of Technology}
  \city{Eindhoven} 
  \state{The Netherlands} 
}
\email{a.yaman@tue.nl}

\author{Decebal Constantin Mocanu}
\affiliation{%
  \institution{Eindhoven University of Technology}
  \city{Eindhoven} 
  \state{The Netherlands} 
}
\email{d.c.mocanu@tue.nl}

\author{Giovanni Iacca}
\affiliation{%
  \institution{University of Trento}
  \city{Trento} 
  \state{Italy} 
}
\email{giovanni.iacca@gmail.com}

\author{George Fletcher}
\affiliation{%
  \institution{Eindhoven University of Technology}
  \city{Eindhoven} 
  \state{The Netherlands} 
}
\email{g.h.l.fletcher@tue.nl}

\author{Mykola Pechenizkiy}
\affiliation{%
  \institution{Eindhoven University of Technology}
  \city{Eindhoven} 
  \state{The Netherlands} 
}
\email{m.pechenizkiy@tue.nl}


\renewcommand{\shortauthors}{A. Yaman et al.}

\begin{abstract}
Many real-world control and classification tasks involve a large number of features. When artificial neural networks (ANNs) are used for modeling these tasks, the network architectures tend to be large. Neuroevolution is an effective approach for optimizing ANNs; however, there are two bottlenecks that make their application challenging in case of high-dimensional networks using direct encoding. First, classic evolutionary algorithms tend not to scale well for searching large parameter spaces; second, the network evaluation over a large number of training instances is in general time-consuming. In this work, we propose an approach called the \textit{Limited Evaluation Cooperative Co-evolutionary Differential Evolution algorithm (LECCDE)} to optimize high-dimensional ANNs.

The proposed method aims to optimize the pre-synaptic weights of each post-synaptic neuron in different subpopulations using a Cooperative Co-evolutionary Differential Evolution algorithm, and employs a limited evaluation scheme where fitness evaluation is performed on a relatively small number of training instances based on fitness inheritance. We test LECCDE on three datasets with various sizes, and our results show that cooperative co-evolution significantly improves the test error comparing to standard Differential Evolution, while the limited evaluation scheme facilitates a significant reduction in computing time.

\end{abstract}

%
%



\maketitle

\section{Introduction}

Scaling artificial neural networks (ANNs) up to solve large complex problems achieved a big success in various machine learning problems. The backpropagation and stochastic gradient descent algorithms are conventional methods for training ANNs~\cite{lecun1998gradient}. An alternative approach, Neuroevolution (NE)~\cite{floreano2008neuroevolution}, employs evolutionary algorithms to optimize the topology and/or weights of the ANNs. The NE algorithms do not require the gradient information, and perform remarkably well in optimizing ANNs based on the direct interaction with their environment; specifically, in the cases where good decision instances are noisy or not known for supervised learning \cite{floreano2008neuroevolution,whitley1993genetic}.

There are mainly two types of NE approaches: direct and indirect encoding \cite{floreano2008neuroevolution}. Direct encoding aims to evolve the network parameters directly representing them within the genotype of the individuals; whereas, indirect encoding aims to evolve the specifications to define the developmental process of an ANN represented within the genotype. The indirect encoding methods can help improving the scalability of the evolutionary process for large networks, since they can reduce the parameter size. On the other hand, the NE with direct encoding presents a challenging opportunity for stimulating the research in large-scale optimization, but also contributes to understanding the evolutionary dynamics of ANNs by suggesting successful evolutionary strategies to evolve ANNs.

The task of evolving direct-encoded large networks is challenging due to 1) the scalability of the evolutionary methods to perform the optimization process efficiently on high-dimensional search spaces, and 2) the time requirement for evaluating the individuals on a large number of training instances. The Cooperative Co-evolution (CC) is an effective approach for optimizing large-scale problems \cite{potter1994cooperative}; and the Limited Evaluation (LE) is an advantageous method for reducing the number of instances of fitness evaluations \cite{morse2016simple}. In this work, we propose a \textit{Limited Evaluation Cooperative Co-evolutionary Differential Evolution} (LECCDE) algorithm that employs the CC and LE approaches to perform accelerated evolution in optimizing high-dimensional ANNs with direct encoding.

With respect to the previous works, the work presented in this paper contributes as follows: 1) it considers the post-synaptic neurons as the building blocks of an ANN, and performs the subcomponent decomposition of the CC scheme by assigning the pre-synaptic weights of each post-synaptic neuron to a subpopulation; 2) it demonstrates the effectiveness of the CC in optimizing large-scale ANNs, and compares with the standard Differential Evolution (DE) optimization; 3) it shows that the LE scheme enhanced with the CC achieves better accuracy results than standard DE for evolving large networks, while reducing the time required for the fitness evaluation.

Three datasets were chosen to evaluate the performance of the proposed algorithm on supervised learning tasks. We used a fully connected feed forward ANNs with one hidden layer, with a total number of parameters in the order of thousands. 
We refer to these ANNs as ``large-scale" in the sense of NE with direct encoding, and to distinguish them from the specialized networks used in Deep Learning (DL) approaches \cite{lecun2015deep}.

The rest of the paper is organized as follows: in Section  \ref{sec:relatedWork}, we provide the background knowledge and a brief literature review on the topics of DE, CC, and NE; in Section \ref{sec:proposedAlgorithm}, we discuss the proposed algorithm in detail; in Section \ref{sec:experimentalSetup}, we present the experimental setup; in Section \ref{sec:results}, we provide the numerical results; and finally, in Section \ref{sec:conclusion}, we discuss the conclusions.

\vspace{-0.2cm}

\section{Related Work} \label{sec:relatedWork}

In this section, we provide a brief overview of the background and related work.

\vspace{-0.2cm}
\subsection{Differential Evolution}

The DE algorithm is a powerful yet simple population-based search algorithm for continuous optimization~\cite{storn1997differential}. A candidate solution set consists of $NP$ individuals represented as $D$-dimensional real-valued vectors $\boldsymbol{x_i} \in \mathbb{R}^{D}$ where integer $i\in [1,NP]$. An initial population of individuals is randomly sampled from the domain ranges of each dimension $x_{i,j}^{min}$ and $x_{i,j}^{max}$ where $x_{i,j}$ is the $j$th dimension of $i$th individual.

In each generation $g$, an individual $\boldsymbol{x_i^g}, \forall i \in (1,2, \cdots, NP)$, called the \textit{target vector}, is selected. The \textit{mutation} and \textit{crossover} operators are applied to generate a \textit{trial vector} $\boldsymbol{u^g}$. The trial vector is evaluated, and replaced with the target vector by the \textit{selection operator}, if the fitness value of the trial vector is greater than or equal to the target vector.

The mutation operator generates a \textit{mutant vector} $\boldsymbol{v_i^g}$ by perturbing a randomly selected vector using the scaled differences of the other two randomly selected vectors. The magnitude of the perturbation is controlled by the parameter called \textit{scale factor} $(F)$. This strategy is referred to as the \textit{``rand/1''} strategy, and is provided by the following equation:
\begin{equation}
\boldsymbol{v_i^g} = \boldsymbol{x_{r_1}^g} + F \cdot ( \boldsymbol{x_{r_2}^g} - \boldsymbol{x_{r_3}^g} ) \label{eq:DErand1}
\end{equation}
\noindent where $r_1, r_2, \mbox{ and } r_3$ are mutually exclusive integers different from $i$, and selected randomly from the range of $[1,NP]$. There are various alternative mutation strategies proposed in the literature \cite{das2016,neri2010}.

The crossover operator recombines the target vector with the mutant vector, controlled by the parameter called \textit{crossover rate} $(CR)$. The \textit{binomial (uniform)} and \textit{exponential} crossover operators are the two most commonly used crossover operators \cite{price2006differential}. The binomial crossover operator is provided in Equation \eqref{eq:uniformCrossover}:
\begin{equation}
u_{i,j}^g = \left\{
  \begin{array}{lr}
    v_{i,j}^g, & \mbox{if } rand([0,1))\leq CR \mbox{ or }j=randi([1,D]);\\
    x_{i,j}^g, & \mbox{otherwise.}
  \end{array}
\right. 
\label{eq:uniformCrossover}
\end{equation}
\noindent where integer $j\in [1,D]$ refers to the $j$th dimension of the vectors, functions $rand()$ and $randi()$ uniformly samples real and integer values within the specified ranges respectively.

The selection operator performs a comparison of the fitness values of the target and trial vectors, and replaces the target vector with the trial vector in the next generation if a better fitness value is achieved by the trial vector. This is referred to as \textit{synchronous} update, since the replacements are performed at the end of the generation when the process for all individuals is complete. The \textit{asynchronous} version of the update is implemented by performing the replacement immediately within the same generation. The asynchronous update allows a newly replaced trial vector to be used by other individuals within the same generation.

The settings of the parameters in DE plays an influential role in the behavior of the algorithm for balancing the trade-off between the exploration and exploitation \cite{neri2010,vcrepinvsek2013exploration}. A recent survey by Karafotias \textit{et al.} reviewed the approaches for parameter tuning and control in evolutionary algorithms \cite{karafotias2015parameter}. Neri and Tirronen surveyed the existing works in the literature and performed an empirical analysis of the strategies and parameters in DE; more recently, Das \textit{et al.} reviewed the works in the literature on the self-adaptive parameter control in DE \cite{das2016}.

\vspace{-0.2cm}

\subsection{Cooperative Co-evolution}

While the dimensionality of a problem increases, the performance of the evolutionary algorithms tend to decrease \cite{mahdavi2015metaheuristics,liu2001scaling}. The CC schemes were proposed for scaling evolutionary algorithms to higher dimensions using a divide-and-conquer strategy. In the CC, the subcomponents of a large-scale problem is decomposed and assigned to a number of subpopulations, that are evolved separately \cite{potter1994cooperative}. Cooperation in co-evolution arises during the fitness evaluation, where the subcomponents are merged together to assign a global fitness score to a candidate solution.

The three aspects that play a key role in CC are \textit{problem decomposition}, \textit{subcomponent evolution}, and \textit{subcomponent co-adaptation} \cite{yang2008large}. The maximum number of subpopulations can be generated by splitting a $D$-dimensional problem into $D$ subgroups, assigning each subcomponent (dimension) to one subpopulation. Alternatively, the number of subcomponents in each subpopulation can be chosen arbitrarily to make the evolutionary optimization process manageable by reducing the dimensionality per subgroup. However, an arbitrary assignment of subcomponents may not be effective for solving non-separable problems. Ideally, the problem should be decomposed in a way that the interdependency between the subcomponents in different subpopulations should be minimized.

The existing knowledge about the problem domain can be beneficial in the problem decomposition process. If the interdependencies of the subcomponents are known, the problem can be decomposed based on this knowledge. This also relates to the separability property of the problem. If the problem is separable, then the problem can be decomposed into its separable subcomponents. If there is no/uncertain knowledge of the problem domain, then automated methods can be used to identify the interactions of the subcomponents \cite{sun2017recursive,omidvar2014cooperative}.

The subcomponent evolution can be performed by using various kinds of evolutionary algorithms \cite{mahdavi2015metaheuristics}, including the DE \cite{shi2005cooperative}. 

\vspace{-0.42cm}

\subsection{Neuroevolution}

ANNs are computational models that are inspired by the central nervous system \cite{de2006fundamentals}. NE is a field that aims to optimize ANNs by using evolutionary computing methods \cite{floreano2008neuroevolution}. The approaches suggested in NE can be grouped as \textit{direct} and \textit{indirect encoding} methods. 
One of the first examples of the direct encoding approaches evolved the connection weights of fixed topology networks by representing them within the genotype of the individual in the population~\cite{whitley1990genetic,fogel1990evolving}.

Neuroevolution of Augmenting Topologies (NEAT) has been proposed to evolve both the topology and the weights of the networks starting from minimal networks and incrementally grow larger networks through the evolutionary process~\cite{stanley2002evolving}. NEAT uses a \textit{global innovation counter} to keep track of the history of changes, and to align the networks to generate more meaningful offspring as a result of the crossover operator.

Some of the works incorporate the CC scheme within Neuroevolution. The Symbiotic Adaptive Neuroevolution (SANE) evolves two separate populations, one for neurons and another for the network ``blueprints". The evolved network blueprints are used to determine which combinations of the neurons to use from the neuron population to generate a network \cite{moriarty1997symbiotic}. The Enforced SubPopulations (ESP) initiates a subpopulation for each neuron, and the genotype of these neurons encode the weights for incoming, outgoing and bias connections \cite{gomez2003robust}; Cooperative Synapse Neuroevolution (CoSyNE) initiates a subpopulation for each connection \cite{gomez2008accelerated}.

The indirect NE methods can help scaling evolutionary approaches for evolving large networks. Kitano \cite{kitano1990designing} suggested a grammatical graph encoding method, based on graph rewriting rules represented as individuals' genotypes, to evolve the connectivity matrix of ANNs. Koutnik \textit{et al.,} proposed using lossy compression techniques to reduce the high-dimensional parameters of the networks by transforming their parameters to the frequency domain using transformation functions such as the Fourier Transform and the Discrete Cosine Transform. In this case the evolutionary process is performed on a few significant coefficients on the frequency domain \cite{koutnik2010evolving}. Gruau suggested a developmental method that evolves tree-structured programs to specify the instructions to grow ANNs based on cell division and differentiation\cite{gruau1994automatic}. Stanley \textit{et al.,} proposed a Hypercube-Based Encoding method that uses Compositional Pattern Producing Networks (CPPNs) to assign the connection weights between neurons as a function of their locations~\cite{stanley2009hypercube,stanley2007compositional}.

The ANN architectures used in DL are often engineered for certain tasks in computer vision and signal processing\cite{lecun2015deep}. In this case the connection weights are typically trained using the backpropagation. On the other hand, there are hyper-parameters for specifying the architecture and learning algorithms that play a role in the performance of network; thus, deep NE approaches have been suggested for optimizing the hyper-parameters of the deep neural networks efficiently \cite{miikkulainen2017evolving,real2017large}. Some recent work focuses on scalable evolutionary approaches for optimizing the connection weights of the networks. Salimans \textit{et al.,} used Evolution Strategies (ES) to optimize the connection weights of the Convolutional Neural Networks (CNNs) for reinforcement learning in MuJoCo and Atari environments \cite{salimans2017evolution}. The CNNs are a specific type of large ANN topologies that are specifically designed for image processing/recognition tasks in DL. Zhang \textit{et al.,} compared the ES proposed by Salimans \textit{et al.} with the stochastic gradient descent for training CNNs on a large handwritten digit dataset, MNIST, and showed that the ES can achieve the state-of-the-art accuracy results~\cite{zhang2017relationship}.

Another scalability challenge for the NE is the fitness evaluation that can be computationally expensive, especially when there are large numbers of training instances to evaluate. Morse and Stanley proposed an approach called \textit{Limited Evaluation Evolutionary Algorithm (LEEA)}, inspired by the batch training in the stochastic gradient descent algorithm. The LEEA performs fitness evaluations over a small number of training instances (batches), and uses accumulated fitness values that are inherited from the parent generation to the offspring generation between batches \cite{morse2016simple}. We adopt the LEEA approach in our algorithm, and discuss the approach in more detail in Section \ref{sec:proposedAlgorithm}.

\vspace{-0.4cm}

\section{The Proposed Algorithm} \label{sec:proposedAlgorithm}

The implementation details of the LECCDE algorithm are given in Algorithm \ref{alg:LECCDE}. The algorithm is composed of the CC and LE schemes to decompose a large-scale continuous optimization task, and speed up the fitness evaluation process.


\begin{algorithm}[!ht]
		\begin{algorithmic}[1]
			\Procedure{LECCDE}{$NP$, $F$, $CR$}
				\State Initialize $NP$ individuals in each subpopulation $P_i, i\in (1,SP)$
               	\State Initialize $Ft_{i,j} \gets 0$ \Comment{ Fitness of the $j$th individual in the $i$th subpopulation}
                \For {$c = 1$ to $trial\times NP$}
                	\State Select a random individual $\boldsymbol{x_{i,{r_j}}}$ from each $P_i$ \Comment {$r_j$ is a randomly generated integer index}
                    \State $\boldsymbol{X} \gets \{\boldsymbol{x_{1,{r_1}}},\boldsymbol{x_{2,{r_2}}},\cdots, \boldsymbol{x_{{SP},{r_{SP}}}}\}$ 
                    \State $Ft_X\gets  evaluate(\boldsymbol{X}, b_1)$ \Comment{$b_1$ is the first batch}
                    \State $\forall \boldsymbol{x_i}\in \boldsymbol{X}, Ft_{i,j}\gets  Ft_{i,j} + Ft_X$
                \EndFor
                \State $\forall i\in (1,SP) \mbox{ and } j\in (1,NP), Ft_{i,j} \gets normalize(Ft_{i,j})$
                \State $\boldsymbol{X} \gets \{\boldsymbol{x_{1,max}},\boldsymbol{x_{2,max}},\cdots, \boldsymbol{x_{SP,max}}\}$ 
                \State {$Ft_{validation} \gets evaluate(\boldsymbol{X}, ValidationSet)$}
                \State {$bestValidation \gets Ft_{validation}$}
                \State {$bestNetwork \gets \boldsymbol{X}$}
                \While{termination criterion is not satisfied}
                	\ForEach {$b_k \in Batches$} \label{lin:batchLoop}
                    	\ForEach {subpopulation $P_i$} \label{lin:subpopLoop}
                        	\State {$P_i^{\prime} \gets P_i$}
                        	\State {$Ft_i^{\prime} \gets Ft_i$}
                        	\ForEach {$\boldsymbol{x_{i,j}} \in P_i$} 
                            	\State $\boldsymbol{X_i} \gets \boldsymbol{x_{i,j}}$
                               	\State $Ft_X \gets evaluate(\boldsymbol{X},b_k)$
                                \State $Ft_X^{\prime} \gets Ft_{i,j}\cdot (1-decay)+Ft_X$ \label{lin:targetFitness}
                                
                                \State $\boldsymbol{v} \gets mutate(\boldsymbol{x_{i,r_1}},\boldsymbol{x_{i,r_2}},\boldsymbol{x_{i,r_3}},F)$ 
                                \State $Ft_v \gets (Ft_{i,r_1} + Ft_{i,r_2} + Ft_{i,r_3}) \diagup 3 $ \label{lin:mutantFitness}
                                \State $\boldsymbol{u} \gets crossover(\boldsymbol{x_{i,j}}, \boldsymbol{v},CR)$
                                \State $\boldsymbol{X_i} \gets \boldsymbol{u}$
                               	\State $Ft_u \gets evaluate(\boldsymbol{X},b_k)$
                                \State $Ft_u^{\prime} \gets ((Ft_{i,j} + Ft_v) \diagup 2) \cdot (1-decay) + Ft_u$ \label{lin:trialFitness}
                                \If {$Ft_u^{\prime} > Ft_X^{\prime}$} 
									\State {$P_{i,j}^{\prime} \gets \boldsymbol{u}$}
                        			\State {$Ft_{i,j}^{\prime} \gets Ft_u^{\prime}$}
                                \Else
                                	\State {$P_{i,j}^{\prime} \gets \boldsymbol{x_{i,j}}$}
                        			\State {$Ft_{i,j}^{\prime} \gets Ft_X^{\prime}$}
								\EndIf 	
                        	\EndFor
                            \State {$P_i \gets P_i^{\prime}$}
                        	\State {$Ft_i \gets Ft_i^{\prime}$}
                            \State {$\boldsymbol{X_i} \gets \boldsymbol{x_{i,max}}$}
                            \State {$Ft_{validation} \gets evaluate(\boldsymbol{X}, ValidationSet)$}
                            \If {$Ft_{validation} > bestValidation$} 
                            	\State $bestNetwork \gets X$								\State $bestValidation \gets Ft_{validation}$
							\EndIf 	
                        \EndFor
                	\EndFor
                \EndWhile
			\EndProcedure
		\end{algorithmic}
		\caption{LECCDE}\label{alg:LECCDE}
	\end{algorithm}

The CC scheme in LECCDE uses a heuristic to decompose the parameters of a high-dimensional ANN, i.e. the post-synaptic neurons are assumed to be the building blocks of the ANN, and are decomposed into subpopulations and evolved separately. Thus, the algorithm initiates $SP$ subpopulations for each post-synaptic neuron, where each subpopulation consists of $NP$ individuals. Each individual represents the pre-synaptic connection weights (see Appx. \ref{apx:neuroevolution}).

From the $SP$ subpopulations that contain $NP$ of individuals, there are $NP^{SP}$ ANNs that can be constructed. To find the average fitness of each individual, all possible network combinations need to be evaluated. Since this number is quite large, we randomly sample $trial \times NP$ times an individual from each subpopulation, construct a global network, evaluate it, and add the fitness value of the network to the fitness values of each individual that was part of the network \cite{gomez2008accelerated}. At the end of this procedure, the fitness value of each individual is normalized to find the average fitness value, dividing by the number of time each individual is selected. The fitness of the individuals that were not selected during the sampling process set to 0. The individual with the maximum fitness from each subpopulation is then selected to construct the global ANN solution $\boldsymbol{X}$. Finally, the performance of the global solution on the validation instances is found by evaluating $\boldsymbol{X}$ on the validation set.

The main loop of the algorithm iterates over all the batches. A \textit{batch} is a small subset of the training instances used in the LE scheme~\cite{morse2016simple}. In particular, $TrainingSize \diagup BatchSize$ batches are generated by randomly assigning each training instance to a batch.

The fitness score of the target vector on the current batch is found by replacing it within its corresponding part in the global solution, and evaluating the global solution on the current batch. Subsequently, the fitness of the target vector is adjusted using the \textit{asexual} reproduction rule (see Appx. \ref{apx:limitedEvaluation}). 

The fitness of the trial vector is computed in a similar fashion, by first replacing its corresponding part within the global solution, and then evaluating the global solution on the current batch. Since the mutant vector is composed of three randomly selected individuals $\{\boldsymbol{x_{i,r_1}},\boldsymbol{x_{i,r_2}}, \boldsymbol{x_{i,r_3}}\}$, the fitness value of the mutant vector is computed by taking their average. 
The fitness value of the trial vector is found using the \textit{sexual} reproduction rule (see Appx. \ref{apx:limitedEvaluation}).

The selection operator copies the trial vector and its fitness to a temporary set if its fitness value is greater than or equal to the fitness value of the target vector; otherwise, the target value and its fitness are copied. After all the computations are completed for all individuals in the subpopulation, the subpopulation is updated simultaneously by copying back the individuals and their fitnesses from the temporary sets.

After each subpopulation update, the individual with the highest fitness value in the subpopulation is copied back to the corresponding part of the global solution $\boldsymbol{X}$. The global solution is evaluated on the validation set, and the one that performed the best is stored and provided as a the final result of the algorithm.

\vspace{-0.25cm}

\section{Experimental setup} \label{sec:experimentalSetup}

Our experimental setup is designed to focus on the following questions:

\begin{enumerate}
\item
Do the ANNs that are evolved using the Cooperative Co-evolutionary DE algorithm with our subpopulation assignment heuristic achieve a better classification accuracy than the ANNs that are evolved by the standard DE algorithm?

\item
Does the LE scheme applied to DE reduce the runtime of the algorithm, without decreasing the classification accuracy of the evolved ANNs?

\end{enumerate}

To answer these questions, we compare the results of the ANNs optimized by four algorithms, DE, LEDE, CCDE, and LECCDE, on three datasets with various sizes. The details for the implementation of the LECCDE are given in Algorithm \ref{alg:LECCDE}. The CCDE and LEDE are implemented in a similar way, but, without the batch loop and the subpopulations, respectively. 
In standard DE, both batch training and subpopulations are not used. The LE algorithms require two evaluation per generation (target and trial vectors are evaluated on the current batch), while the algorithms without LE require one evaluation per generation. Regardless of this fact, the algorithms were run for the same number of function evaluations (FEs) for each dataset. For all experiments, we used \textit{``rand/1/bin"} (\textit{``rand/1"} mutation with \textit{binomial} crossover) strategy with empirically fixed the parameter settings of $F$ and $CR$ to $0.1$ and $0.3$, respectively. We used $20$ individuals for the population size, except for one experiment that we performed on a larger population size consisting of $100$ individuals (see below). We set $trial$ parameter to 5.

The three datasets used in the test process are listed in Table \ref{tab:datasets}. These datasets were obtained from the Center for Machine Learning and Intelligent Systems dataset repository \cite{Lichman2013}. These datasets were chosen based on their number of features and instances, to show the relative performance of the algorithms in respect to the size of the dataset used. The Wisconsin breast cancer (WBC) dataset consists of 30 features, 2 classes, and 569 instances, the epileptic seizure recognition (ESR) consists of 178 features, 2 classes, and 4600 instances\footnote{The original epileptic seizure recognition dataset \cite{andrzejak2001indications} consists of 5 classes (first class for the measurements of the patients who had epileptic seizure and the remaining 4 classes for the measurements of the patients who did not have epileptic seizure), and 11500 instances (2300 for each class). To reduce the complexity of the task we took only the instances from the first and second classes with 2300 instances from each, thus considering 4600 instances in total.}, and the human activity recognition (HAR) dataset consists of 561 features, 6 classes, and 7144 instances \cite{anguita2013public}. The instances in each dataset were split into three groups (training, validation, and test) with ratios $70\%$, $15\%$, and $15\%$ respectively. The fitness evaluations and selection process were performed on the train instances. The network that performs the best on the validation set is provided as the output of the algorithm, and evaluated on the test set. The fitness evaluation is based on the classification accuracy of the ANNs which is calculated by the number of correctly classified instances divided by the total number of instances.


For all datasets, we used fixed-topology fully-connected feed forward ANNs with one hidden layer to perform the classification task (see Appx. \ref{apx:neuroevolution}). The number of neurons within the hidden layer was kept constant at $50$ for all ANNs evolved for all datasets. Based on the architecture of the ANNs and the number of features in the datasets, the total number of parameters evolved are 1652, 9052, and 28406 for the WBC, ESR, and HAR respectively. 

We used a batch size of 100 instances for the WBC, 500 for the ESR, and 500 for the HAR. The decay value (see Appx. \ref{apx:limitedEvaluation}) is set to 0.2, as suggested by Morse and Stanley \cite{morse2016simple}. The maximum number of FEs was set to 50000 for the WBC, 300000 for the ESR, and 500000 for the HAR, based on the number of their parameters.

\begin{table}[]
\small
\centering
\caption{The specifications of the datasets used in the experiments.}
\label{tab:datasets}
\begin{tabular}{|l|l|l|l|l|}
\hline
\textbf{Datasets} & \textbf{Features} & \textbf{Classes} & \textbf{Instances} & \textbf{Parameters} \\ \hline
WBC              & 30                     & 2                     & 569                     & 1652                       \\ \hline
ESR              & 178                    & 2                     & 4600                    & 9052                       \\ \hline
HAR              & 561                    & 6                     & 7144                    & 28406                      \\ \hline
\end{tabular}
\end{table}


\section{Numerical results} \label{sec:results}

In this section, we present our experimental results. Each algorithm, with the specified settings, was run for 20 independent runs, and the median and the variance of train, validation and test accuracy were collected. All the accuracy results are shown with a precision of two digits.

Table \ref{tab:wbcResults} shows the results obtained from the WBC dataset. In this case we could not observe a significant difference on the results of the ANNs evolved by the four algorithms. On the test data, the CCDE appears to be performing better than others. On the other hand, we observe a difference on the runtime of the algorithm ($t=322$ sec, in our computing environment\footnote{All algorithms were run, in single-core, on an Intel Xeon E5 3.5GHz computer.}). The algorithms that employ LE and CC are less computationally expensive and run faster. For example, the runtime of DE is more than twice as big as that of LECCDE. This difference is less significant for the other algorithms, due to the size of the dataset. Even though all the algorithms are run for 50000 FEs for this dataset, the algorithms with LE performed evaluation on batches that are four times smaller than the whole set of training instances. However, since CCDE is run on the whole dataset, it appears that the CC improved its runtime possibly due to the computations of reduced-sized vectors within each subpopulation.

\begin{table}[]
\small
\centering
\caption{The median of the accuracy results of the ANNs evolved using four variants of DEs on the WBC dataset.}
\label{tab:wbcResults}
\begin{tabular}{|l|l|l|l|l|}
\hline
\textbf{Alg.} & \textbf{Train}  & \textbf{Validation} & \textbf{Test}   & \textbf{Runtime} \\ \hline
\textbf{DE}        & 94.74 $\pm$ 2.2  & 97.65 $\pm$ 0.8     & 95.29 $\pm$ 2.2 & $2.12\times t$   \\ \hline
\textbf{LEDE}      & 95.99 $\pm$ 1.2 & 98.82  $\pm$ 0.6    & 95.29 $\pm$ 2.3  & $1.43 \times t$   \\ \hline
\textbf{CCDE}      & 96.49 $\pm$ 1.3  & 97.65 $\pm$ 0.8     & 96.47 $\pm$ 1.8 & $1.10 \times t$   \\ \hline
\textbf{LECCDE}    & 96.24 $\pm$ 1.9 & 97.65 $\pm$ 0.8     & 95.29 $\pm$ 2.8 & $t$       \\ \hline
\end{tabular}
\end{table}

Table \ref{tab:ESRresults} presents the results obtained from the ESR dataset. Based on the test data, CCDE appears to show better performance than the rest of the algorithms, while LECCDE follows it very closely. We observe the best running time with LECCDE ($t = 2970$ sec). 

\begin{table}[]
\small
\centering
\caption{The median of the accuracy results of the ANNs evolved using four variants of DEs on the ESR dataset.}
\label{tab:ESRresults}
\begin{tabular}{|l|l|l|l|l|}
\hline
\textbf{Alg.} & \textbf{Train} & \textbf{Validation} & \textbf{Test}   & \textbf{Runtime} \\ \hline
\textbf{DE}        & 90.50 $\pm$ 1.3         & 89.86 $\pm$ 1.2            & 89.57 $\pm$ 1.7         & $2.66\times t$            \\ \hline
\textbf{LEDE}      & 92.86 $\pm$ 0.9         & 92.25 $\pm$ 0.8              & 91.30 $\pm$ 0.9          & $1.26\times t$           \\ \hline
\textbf{CCDE}      & 93.94  $\pm$ 0.8        & 93.33 $\pm$ 0.3             & 92.17 $\pm$ 0.9 & $2.05\times t$            \\ \hline
\textbf{LECCDE}    & 93.98 $\pm$ 0.6         & 92.61 $\pm$ 0.5              & 91.88 $\pm$ 1.0 & $t$                \\ \hline
\end{tabular}
\end{table}

Table \ref{tab:HARResults} shows the results obtained from the HAR dataset. On this dataset, we observe a significant accuracy improvement when the CC scheme is used. The performance of CCDE and LECCDE are approximately \%15-20 better than the algorithms that do not use the CC. Also, CCDE appears to be slightly better than LECCDE. On the other hand, we observe a significant runtime improvement when the LE scheme is used. The algorithms with the LE scheme run approximately four times faster than the algorithms that do not use the LE ($t = 6530$ sec). Also, LECCDE appears to produce the smallest variance on th train accuracy.

\begin{table}[]
\small
\centering
\caption{The median of the accuracy results of the ANNs evolved using four variants of DEs on the HAR dataset.}
\label{tab:HARResults}
\begin{tabular}{|l|l|l|l|l|}
\hline
\textbf{Alg.} & \textbf{Train} & \textbf{Validation} & \textbf{Test}   & \textbf{Runtime} \\ \hline
\textbf{DE}        & 70.06 $\pm$ 2.9         & 70.06 $\pm$ 2.7              & 68.38 $\pm$ 3.0          & $4.75\times t$             \\ \hline
\textbf{LEDE}      & 77.5 $\pm$ 5.2         & 77.99 $\pm$ 4.8              & 76.96 $\pm$ 4.8          & $1.28\times t$             \\ \hline
\textbf{CCDE}      & 94.01 $\pm$  0.8       & 92.72 $\pm$ 0.7              & 92.4 $\pm$ 1.0          & $4\times t$             \\ \hline
\textbf{LECCDE}    & 93.58 $\pm$ 0.6         & 93.19 $\pm$ 0.5               & 92.16 $\pm$ 0.7 & $t$               \\ \hline
\end{tabular}
\end{table}

Finally, in Table \ref{tab:ESRpop100}, we report an additional experiment on the population size. In this case, we used a population size of 100 on the ESR dataset. When the population size increases (comparing to the Table \ref{tab:ESRresults}), the accuracy results decrease. This may be due to the number of FEs needed for the convergence of the algorithm: in other words, when the population increases, the number of FEs needed for the convergence may increase. Moreover, we observe that CCDE and LECCDE perform significantly better than DE and LEDE. This may suggest that the CC increased the convergence speed. With respect to the running time of the algorithms, we observe the similar pattern observed in Table \ref{tab:ESRresults} ($t = 2640$ sec).

Overall, CCDE appears to perform better than LECCDE due to the fact that it has the complete information for evaluating the individuals since it uses the entire set of training instances. However, CCDE comes with a larger runtime trade-off than LECCDE, which can make the difference with large datasets (e.g. for the HAR dataset the LECCDE runs on average four times faster). Also, increasing the number of evaluations or batch size can improve the performance of the LECCDE. For comparison, we performed two additional experiments with LECCDE, with the same settings used to produce the results in the Table \ref{tab:HARResults}, except the number of FEs and batch size. In the first experiment, we used 900000 FEs and observed that the ANNs the LECCDE optimize perform on training, validation and test sets on average 95.78, 94.31, and 93.28 respectively. This is almost \%1 higher than  the the performance observed in \ref{tab:HARResults}. On the other hand, the runtime of the algorithm is now $1.6 \times t$, which is still $1.8$ times faster than the runtime of CCDE. In the second experiment, we used a batch size of 1000, and we observed that the algorithm performs on average 96.60, 93.84, and 93.38 on training, validation, and test datasets, with a runtime of $1.42 \times t$. These two additional experiments show an interesting trade-off between the batch size and the number of evaluations. Although the two additional experiments have similar runtime, the second experiment appears to produce better results. 

\begin{table}[]
\small
\centering
\caption{The median of the accuracy results of the ANNs evolved using four variants of DEs on the ESR dataset using population size of 100.}
\label{tab:ESRpop100}
\begin{tabular}{|l|l|l|l|l|}
\hline
\textbf{Alg.} & \textbf{Train} & \textbf{Validation} & \textbf{Test}   & \textbf{Runtime} \\ \hline
\textbf{DE}        & 81.99 $\pm$ 1.4         & 82.61 $\pm$ 1.2              & 80.65  $\pm$ 1.9        & $2.62\times t$              \\ \hline
\textbf{LEDE}      & 80.25 $\pm$ 1.4         & 81.59 $\pm$ 0.8              & 80.14  $\pm$ 2.2        & $1.30\times t$             \\ \hline
\textbf{CCDE}      & 91.65 $\pm$ 1.3        & 91.45 $\pm$ 0.7              & 90.29 $\pm$ 0.9          & $2.01\times t$             \\ \hline
\textbf{LECCDE}    & 91.27 $\pm$   0.8      & 90.58 $\pm$ 0.7              & 88.99 $\pm$ 1.1 & $t$                \\ \hline
\end{tabular}
\end{table}

\begin{figure}[h]
    \centering
    \includegraphics[width=0.4\textwidth]{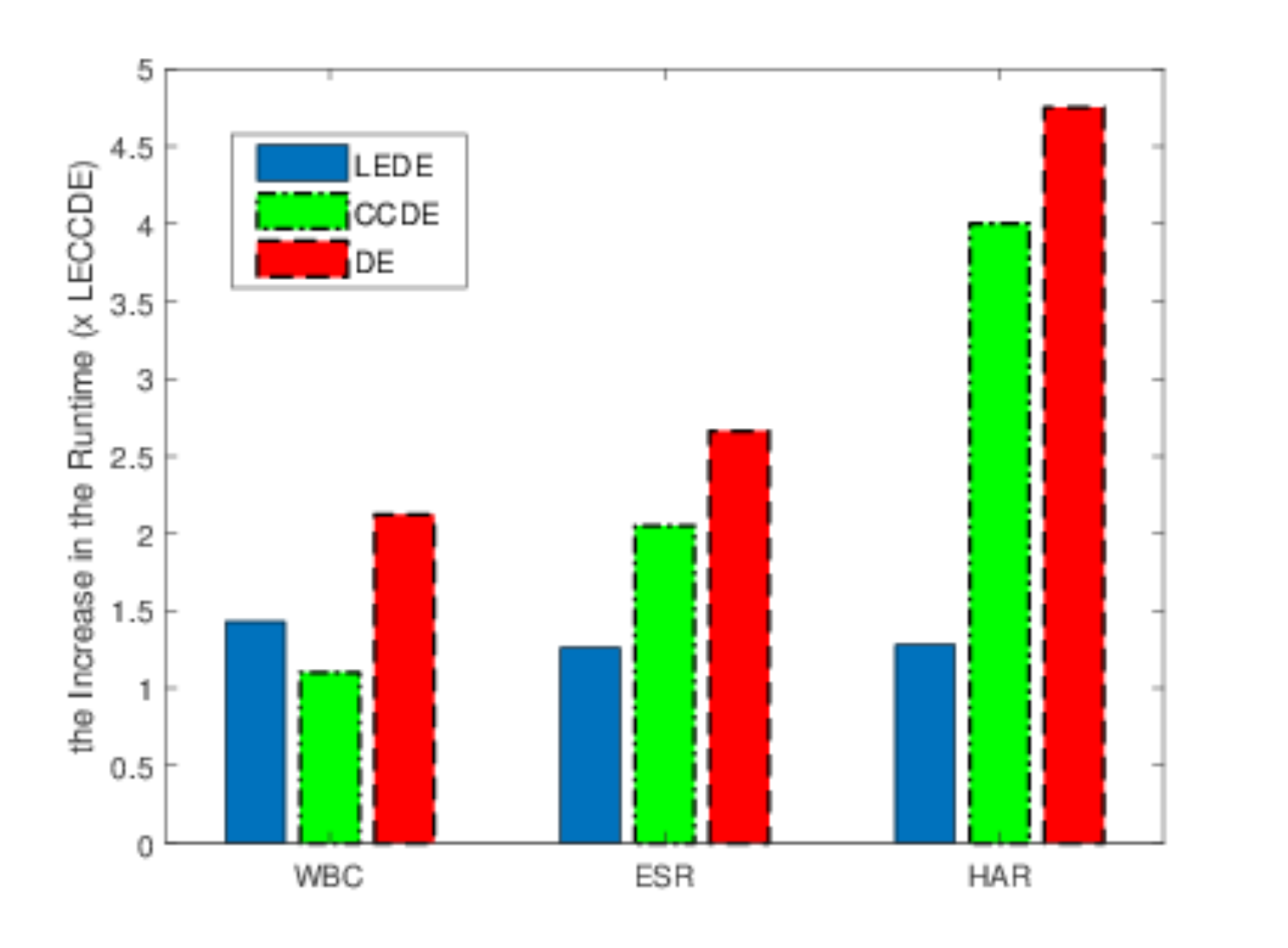}
    \caption{The increase in the runtime of each algorithm relative to the LECCDE on the three datasets. }
    \label{fig:figureCompare}
\end{figure}

Figure \ref{fig:figureCompare} shows the overall comparison of the runtime of LEDE, CCDE, and DE relative to LECCDE on the three datasets. The $x$-axis shows the dataset, and the $y$-axis shows the increase in the runtime of the algorithm. The LEDE is relatively stable across experimented datasets. On the other hand, the runtime of the CCDE and DE increases when the number of instances increases. This is because the algorithms with LE perform the same number of function evaluations, on a smaller number of instances, which produces a clear advantage in terms of total runtime.

\begin{figure}[h]
    \centering
    \includegraphics[width=0.4\textwidth]{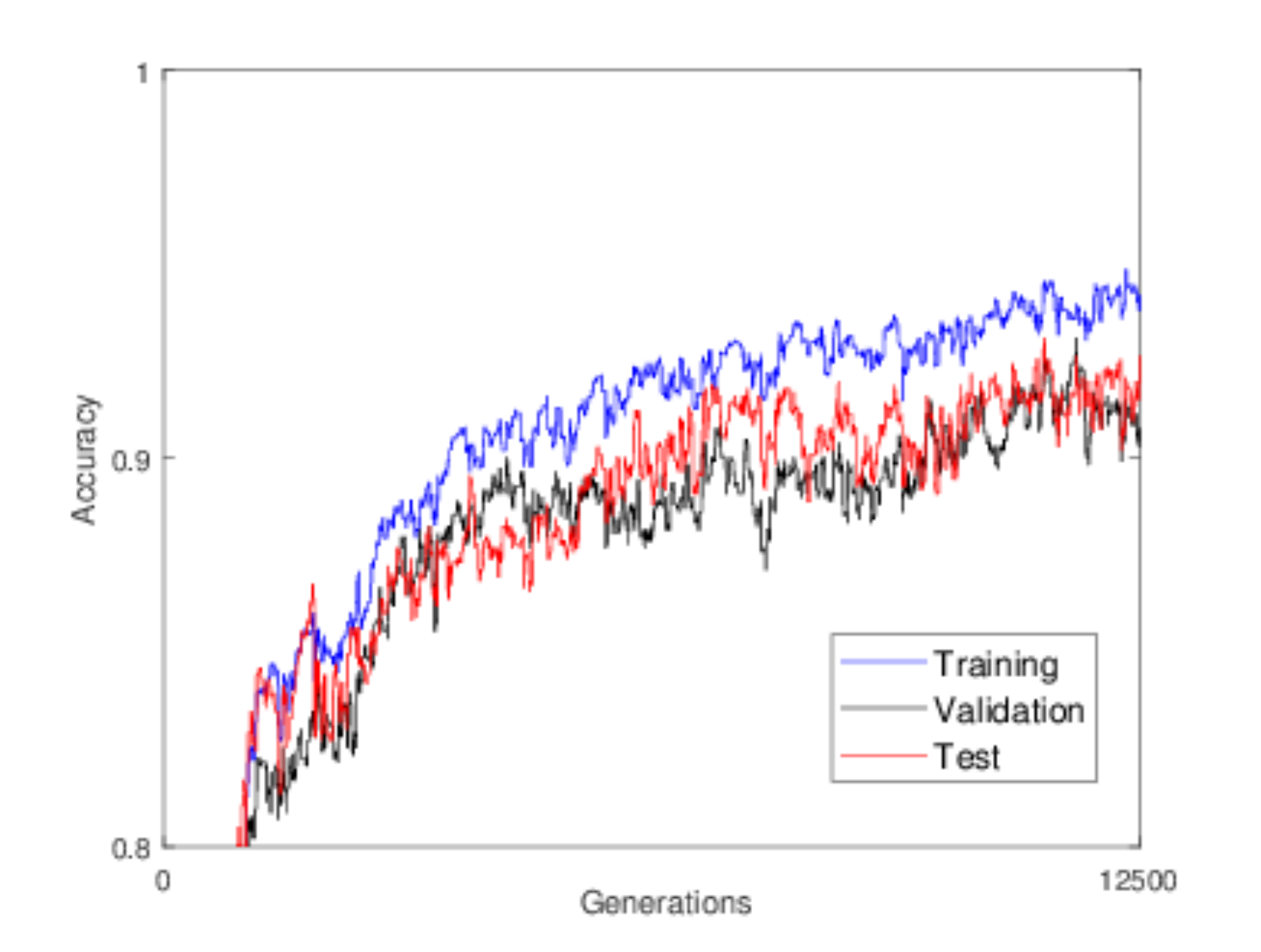}
    \caption{The change of the accuracy results of the ANNs on the training, validation, and test instances while the LECCDE algorithm is running. }
    \label{fig:evoRun}
\end{figure}

Figure \ref{fig:evoRun} shows the accuracy trend of the ANNs on the training, validation, and test instances during one example run of the optimization process performed by LECCDE (only the range $[0.8,1]$ is shown on the $y$-axis, for the sake of clarity). The data collected from this specific run shows that the accuracy on the training data is almost always the highest. The accuracy results of the test data closely follows the validation accuracy, and it is even higher for some of generations.

\begin{figure}[h]
    \centering
    \includegraphics[width=0.4\textwidth]{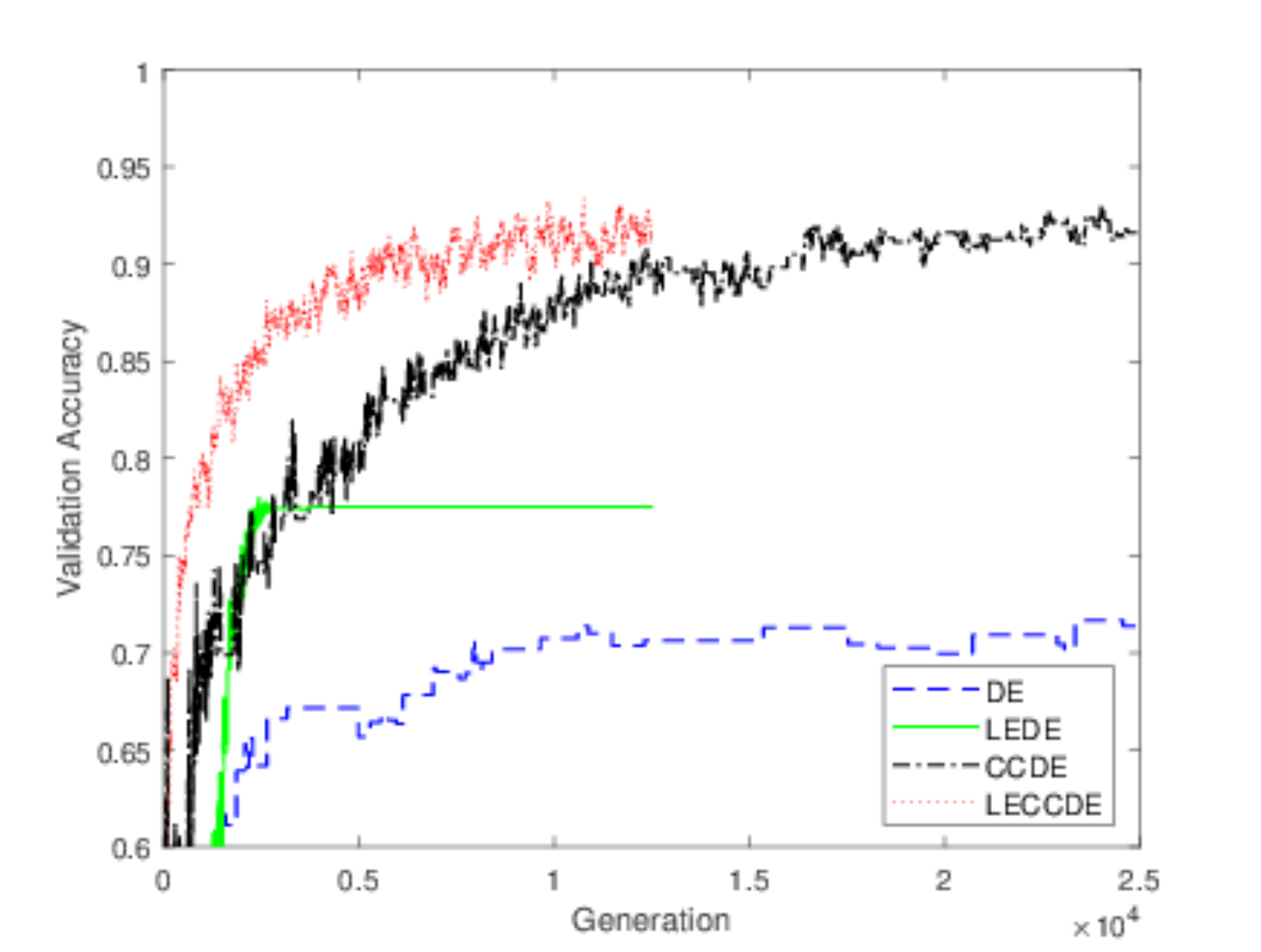}
    \caption{A single run of the change of the validation accuracy during the evolutionary process of four algorithms on the HAR dataset. }
    \label{fig:runtimeComparison}
\end{figure}

Figure \ref{fig:runtimeComparison} shows the change of the validation accuracy during the evolutionary process of four algorithms on a single run (only the range $[0.6-1]$ is shown on the $y$-axis). Firstly, the lines that represent the results of the LEDE and LECCDE are shorter than those of the other algorithms because they consume the same number of FEs within a half number of generations, since they perform two FEs (trial and target vectors) per generation. We observe that LEDE improves the DE in terms of validation accuracy and convergence speed; however, it suffers from the lack of diversity within the population (for a population size of 20), which prevents it from finding better solutions after about 80000 FEs are consumed. On the other hand, CCDE appears not to suffer from the early convergence problem observed in the LEDE, while LECCDE appears to improve the speed of CCDE.

To summarize, our empirical analysis suggests positive answers to the questions posed in Section \ref{sec:experimentalSetup}: (1) It appears more significantly on large dataset (in Table \ref{tab:HARResults}), or with a large population size (in Table \ref{tab:ESRpop100}), that the ANNs that are evolved using the CC scheme using our heuristic achieve a better classification accuracy than the ANNs that are evolved by the standard DE algorithm; and (2) all experiments on the three datasets (most significantly on the largest dataset in Table \ref{tab:HARResults}), show that the LE scheme applied to DE reduce the runtime of the algorithm considerably, without causing a degradation on the classification accuracy of the evolved ANNs. 

To further assess the scalability of the proposed algorithm, we performed an additional experiment on the MNIST dataset~\cite{lecun1998gradient}. We used the same ANN architecture that was used in the previous experiments. We provide the numerical results ---which are not shown here for brevity--- on the extended version of the paper available online\footnote{Supplementary results available at: https://arxiv.org/abs/1804.07234}. For the same number of function evaluations, the computing time required for the LEDE and LECCDE is about 25 times less than the computing time required for the DE and CCDE. The LECCDE performs 8\% better than LEDE. Overall, our preliminary results on MNIST show that the LECCDE achieves 90.80\% classification accuracy on the test data, on average, which is about 4\% lower than the backpropagation algorithm on the same ANN architecture. This may suggest that a better parameter tuning may be needed for the LECCDE to obtain results which are comparable to the state-of-the-art.

\section{Conclusions} \label{sec:conclusion}
In this work, we proposed the LECCDE algorithm that employs the LE and CC schemes to improve the accuracy and the runtime of the standard DE algorithm for large-scale NE with direct encoding.

We performed experiments on four datasets, including a preliminary test on the MNIST dataset. Our results show that the CC scheme improves the performance of DE on the tested classification tasks. Moreover, we used the LE scheme to further improve the scalability of the method. Our results show that the LE scheme reduces the runtime of the algorithms, without affecting the performance. This reduction is due to the fact that the evaluation is performed on a small number of instances.

We used a heuristic in the CC scheme that decomposes the problem at the level of post-synaptic neurons. Thus, we evolve all the pre-synaptic weights of the post-synaptic neurons in different subpopulations. This decomposition approach aims to reduce the parameter size per subpopulation. For large datasets on the other hand, the number of parameters per subpopulation may still be large. Although this heuristic worked well, there may also be other decomposition heuristics that can be more effective. Alternatively, automatic methods can also be used for this purpose.


Another possibility for improving the results can be achieved by performing a sensitivity analysis. In this work, we did not experiment on the strategy and the parameters settings of the DE algorithm. Self-adaptive parameter control approaches can also be investigated to improve the performance of the results since these approaches can adjust the balance between the exploration and exploitation during the search process \cite{yaman2018multi,das2016}.  


The methods proposed here can evolve only the ANNs with fixed topologies, it will be useful to extend these methods also to the network topology optimization.



\appendix
\section{Neuroevolution} \label{apx:neuroevolution}
\subsection{Direct Encoding and Network Computation}

An example of a feed forward network (FFN) is shown in Figure \ref{fig:FFN} where each node represents a neuron, and each edge represents a connection between two node, and the direction of each edge represents the direction of the information flow. A FFN consists of a number of \textit{input} ($i_1,i_2,i_3, b_1$), \textit{hidden} ($h_1,h_2, b_2$), and \textit{output} ($o_1$) neurons ($b_1$ and $b_2$ are bias neurons kept constant at 1) structured as input, hidden and output layers respectively (see Figure \ref{fig:FFN}). Inspired by biological neural networks, the connections between the neurons are often called \textit{synapses}. A neuron that is at the starting point of the directional edge is called a \textit{pre-synaptic neuron}, and the neuron that is at the end point (arrow) of the directional edge is called a \textit{post-synaptic neuron}.

\begin{figure}[!ht]
\centering
\begin{tabular}{ p{8cm}}

\centering
 \subfloat[]{ \label{fig:FFN}
\begin{tikzpicture}[shorten >=1pt, auto, thick]
    \node[gexamples_node_style,fill=gray!5!] (i1) at (0,0) {$i_1$};
    \node[gexamples_node_style,fill=gray!5!] (i2) at (1.5,0) {$i_2$};
    \node[gexamples_node_style,fill=gray!5!] (i3) at (3,0) {$i_3$};
    \node[gexamples_node_style,fill=gray!50!] (h1) at  (0.75,1.5){$h_1$};
    \node[gexamples_node_style,fill=gray!50!] (h2) at (2.25,1.5) {$h_2$};
        \node[gexamples_node_style,fill=gray!50!] (o1) at (2.25,3) {$o_1$};
        \node[] (b1) at (4.5,0) {$b_1$};
        \node[] (b2) at (3.75,1.5) {$b_2$};
    
    \begin{scope}[every node/.style={scale=0.9}]
    \draw[-stealth]  (i1) edge node[left]{$0.7$} (h1);
    \draw[-stealth]  (i1) edge node[pos=0.1, right]{$2.1$} (h2);
    \draw[-stealth]  (i2) edge node[pos=0.2, left]{$0.8$} (h1);
    \draw[-stealth]  (i2) edge node[pos=0.2, right]{$0.6$} (h2);
    \draw[-stealth]  (i3) edge node[pos=0.1, left]{$0.1$} (h1);
    \draw[-stealth]  (i3) edge node[pos=0.15, right]{$1.2$} (h2);
     \draw[-stealth]  (h1) edge node[left, near start]{$0.3$} (o1);
     \draw[-stealth]  (h2) edge node[right, near start]{$0.5$} (o1);
     \draw[-stealth]  (b2) edge node[right, near start]{$1.3$} (o1);
      \draw[-stealth]  (b1) edge node[left, pos=0.1]{$0.4$} (h1);
      \draw[-stealth]  (b1) edge node[right, near start]{$1.4$} (h2);
    \end{scope}
	\end{tikzpicture}
} \\

\centering
 \subfloat[]{\label{tab:FFNGenotype}
\begin{tabular}
{|p{1.1cm}||p{0.25cm}|p{0.25cm}|p{0.25cm}|p{0.25cm}||p{0.25cm}|p{0.25cm}|p{0.25cm}|p{0.25cm}||p{0.25cm}|p{0.25cm}|p{0.25cm}|}
\hline
\footnotesize \textbf{post-synaptic neurons:} &\multicolumn{4}{|l||}{\textbf{$h_1$}}    &  \multicolumn{4}{l||}{\textbf{$h_2$}}    & \multicolumn{3}{l|}{\textbf{$o_1$}} \\ \hline\hline

\footnotesize \textbf{pre-synaptic weights:} &0.7 & 0.8 & 0.1 & 0.4 & 2.1& 0.6 & 1.2 & 1.4 & 0.3 & 0.5 & 1.3 \\ \hline

\end{tabular}

}
\\

\end{tabular} 

\caption{\protect\subref{fig:FFN} A fully-connected feed-forward ANN with one hidden layer, and \protect\subref{tab:FFNGenotype} the representation of its genotype.}\label{fig:FFNExample} 
\end{figure}

Figure \ref{tab:FFNGenotype} shows the genotype representation of the network given in Figure \ref{fig:FFN}. Each synaptic weight is mapped directly to a gene in the genotype. The genotype is divided into its subcomponents consisting of the pre-synaptic weights of each post-synaptic neuron.

The activation of each neuron is updated using Equation~\eqref{eq:activationPostSynaptic} where $a_i$ is the activation of a post-synaptic neuron, $a_j$ is the activation of the $j$th pre-synaptic neuron and $w_{i,j}$ is the connection between them, $b_i$ is the bias of the post-synaptic neuron, and $\psi$ is an activation function given in \eqref{eq:activation} \cite{de2006fundamentals}.

\begin{equation}
a_i = \psi 
\left( 
\sum_j  w_{i,j} \cdot a_j + b_i
\right) 
\label{eq:activationPostSynaptic}
\end{equation}

\begin{equation}
\psi(x) = \frac{2}{1+e^{-2x}}-1\label{eq:activation}
\end{equation}

\noindent  

\subsection{Limited Evaluation}\label{apx:limitedEvaluation}

When the evaluation is performed episodically on a small subset of the whole training instances (batches), it is required to keep track of the individuals that performed well on the previous episodes. The LE scheme aims to adjust the fitness of the offspring by taking into account the success of its parents by fitness inheritance. The \textit{sexual} and \textit{asexual} reproduction rules are provided in Equations \eqref{eq:asexual} and \eqref{eq:sexual} respectively \cite{morse2016simple}.

\begin{equation}
f^{\prime} = f_{parent} \cdot (1- decay) + f\label{eq:asexual}
\end{equation}

\begin{equation}
f^{\prime} = \frac{f_{parent_1} + f_{parent_2}}{2} \cdot (1-decay) + f \label{eq:sexual}
\end{equation}

{\noindent where, $f^{\prime}$ is the adjusted fitness of the offspring, $f_{parent}$ is the parent of its parent, $f$ is the actual fitness of the offspring on current batch of the training instances, and $decay$ is a constant value for adjusting the weight of the previous fitness evaluations. The sexual reproduction method consists of two parents. 
}

\begin{acks}

\noindent
      \begin{tabular}{p{0.15\linewidth} p{0.8\linewidth}}
      \raisebox{-0.8cm}{\includegraphics[scale=0.14,bb=0 0 100 100]{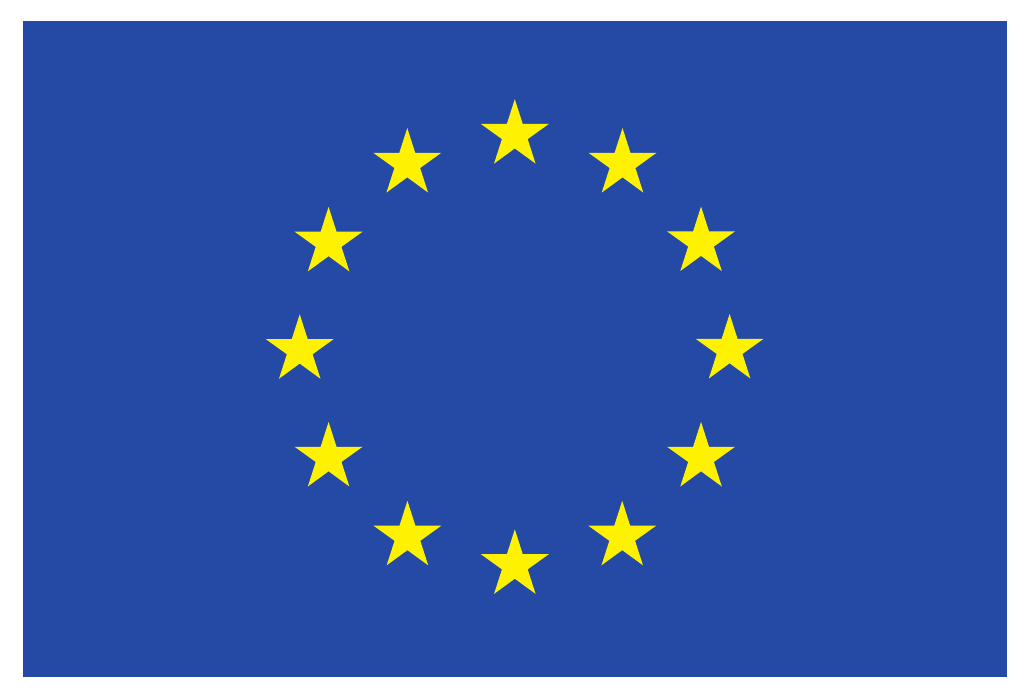}}
      &
  	{\small\fontfamily{arial}\selectfont This project has received funding from the European Union's Horizon 2020 research and innovation programme under grant agreement No: 665347.}
  	\end{tabular}

\end{acks}


\pagebreak

\section{Extended experiments and results}

This section presents our preliminary results of the experiments performed on the MNIST dataset using the DE, LEDE, CCDE and LECCDE. The MNIST dataset consists of 60000 samples of 28 by 28 grayscale image instances of handwritten numbers between 0-9. Thus, the size of the input and output are 784 and 10 when each image pixel and its class label are considered as an input and output respectively. 

We used the same architecture of the artificial neural networks that were used for the experiments performed on the other datasets (feed forward artificial neural networks with one hidden layer consisting of 50 neurons). Thus, the total number of parameters of the networks optimized for the MNIST is 47710. The parameters of the Differential Evolution algorithm are also initialized using the same settings used for the other experiments except for batch size, number of individuals in each subpopulation and the maximum number of function evaluations. Since MNIST is larger than the tested other datasets, we used a a batch size of 1000, a population size of 60 and a maximum number of evaluations set to $2.16e+6$.

Table \ref{tab:MNIST} shows the training, validation and test accuracy results of the ANNs trained for the MNIST dataset. Each variant of the algorithm was executed for the same number of function evaluations. The total time required for computing every other algorithm is shown in relation to the computing time required for the LECCDE where $t = 6.6e+5$ seconds that is approximately 19 hours on a single-core Intel Xeon E5 3.5GHz computer. Due to time constraints, we were able to perform 3 independent runs for the LEDE and LECCDE, and a single partial run for the DE and CCDE. Thus, on DE and CCDE we report their accuracy at 12\% of their total allocated computing time (the total computing times of DE and CCDE are estimated based on their current execution progress).

\begin{table}[!ht]
\footnotesize
\centering
\caption{The accuracy the ANNs evolved for the MNIST dataset, and the runtime of the algorithms.}
\label{tab:MNIST}
\begin{tabular}{|l|l|l|l|l|}
\hline
\textbf{Alg.} & \textbf{Train} & \textbf{Validation} & \textbf{Test}   & \textbf{Runtime} \\ \hline
\textbf{DE} (\% 12)       &   61.60     &  61.26                &   62.52      & $27.2\times t$              \\ \hline
\textbf{LEDE}   & 82.68 $\pm$ 0.36 &  82.01 $\pm$ 0.75  & 82.23  $\pm 0.25$                   & $1.1\times t$             \\ \hline
\textbf{CCDE}   (\% 12)   &  62.40       &  61.80               &  63.20          & $25.3\times t$             \\ \hline
\textbf{LECCDE}    & 91.79 $\pm $ 0.28 &  91.01 $\pm$ 0.63  & 90.80 $\pm$ 0.15                   & $t$             \\ \hline
\end{tabular}
\end{table}

We observe a significant advantage in using the LE scheme on MNIST from the computing time point of view: indeed, the DE and CCDE implementations of the algorithm require a computing time that is 25 times bigger than the computing time required by the corresponding algorithms that make use of the LE scheme.

\begin{figure}[!ht]
    \centering
    \includegraphics[width=0.45\textwidth]{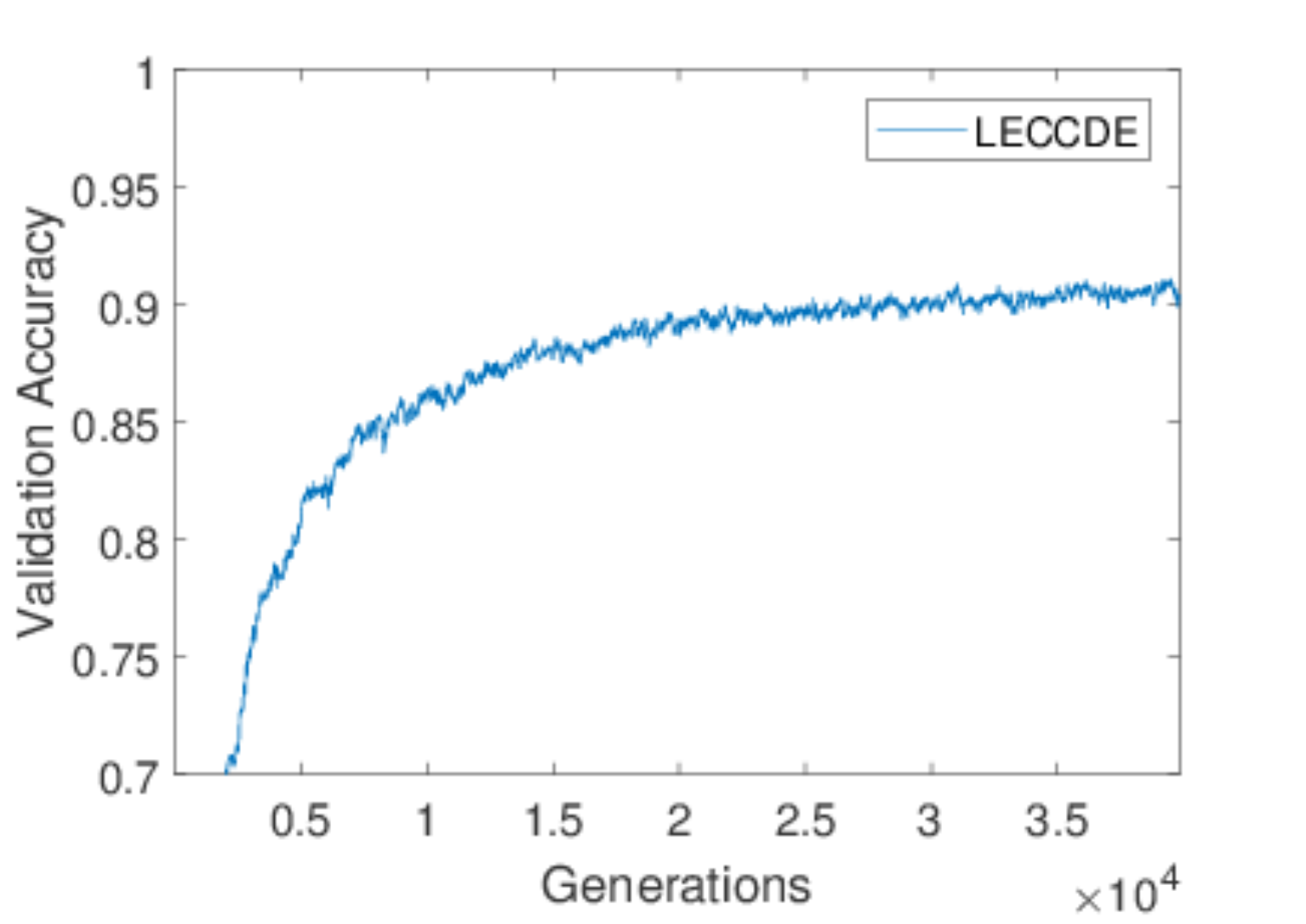}
    \caption{The change of the validation accuracy of the ANNs evolved using the LECCDE on MNIST dataset (only $[0.7,1]$ range is shown on the $y$-axis).}
    \label{fig:evoRunMNIST}
\end{figure}

Figure \ref{fig:evoRunMNIST} illustrates the change of the validation accuracy of the evolved ANNs using the LECCDE during an evolutionary process.  The speed of the accuracy improvements slows down around 88\% - 90\% level. The best validation accuracy achieved during this evolutionary run was 91.62\%.

\end{document}